\documentclass[letterpaper, 10pt, conference]{ieeeconf}      

\IEEEoverridecommandlockouts 
\usepackage{cite}

\usepackage{amsmath,amssymb,amsfonts}
\usepackage{graphicx}
\usepackage{textcomp}
\usepackage[table,xcdraw]{xcolor}
\usepackage{authblk}
\let\labelindent\relax
\usepackage{enumitem}
\usepackage{hyperref}
\usepackage{breqn}
\usepackage{algpseudocode}
\usepackage[linesnumbered,ruled,lined]{algorithm2e}
\usepackage{subfigure}

\usepackage{array}
\usepackage{makecell}
\usepackage{tabularray}

\usepackage{svg}
\usepackage{multirow}

\usepackage{caption}

\allowdisplaybreaks

\makeatletter
\apptocmd{\@maketitle}{
    \centering
    
    \includegraphics[scale=0.23]{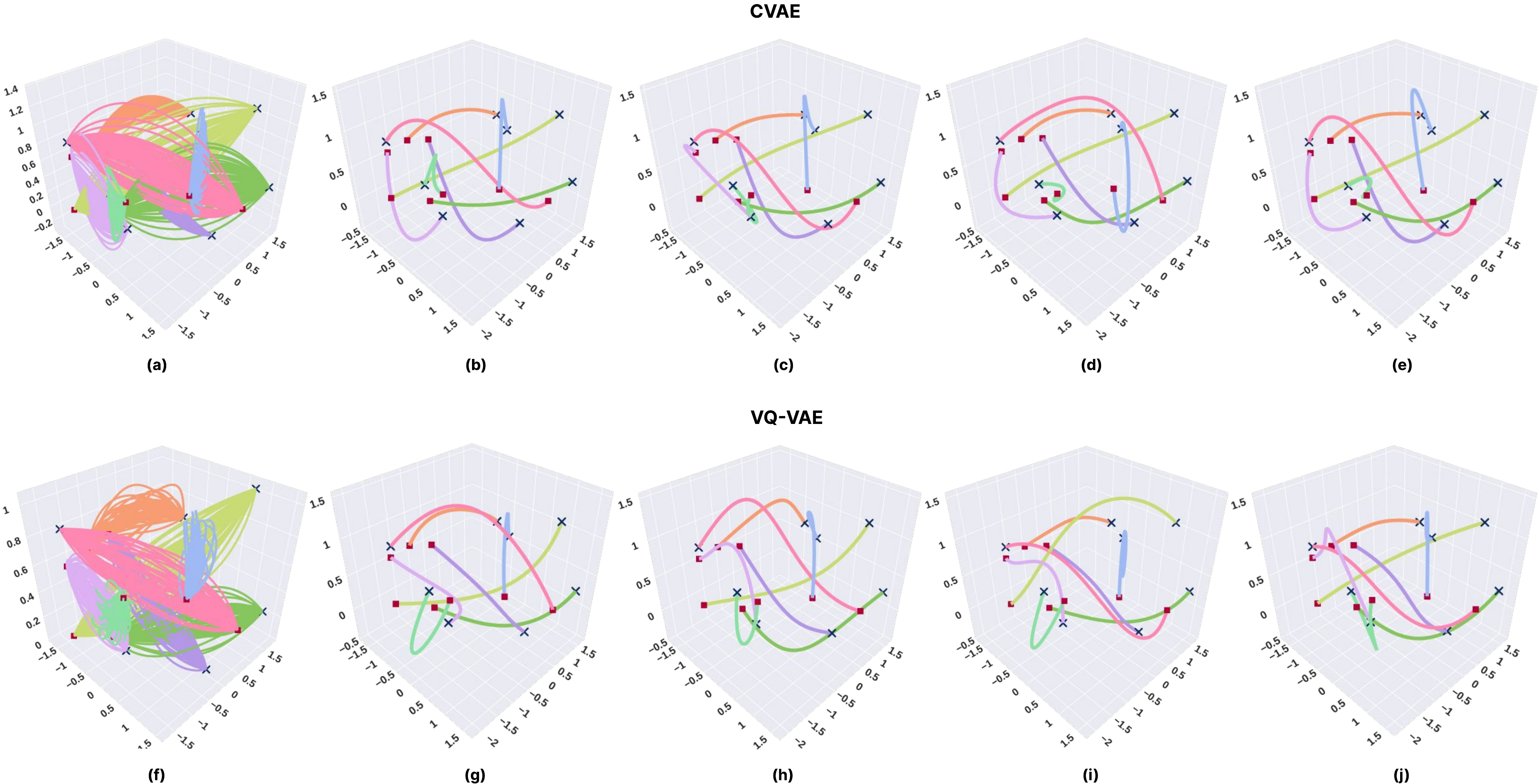}
    \captionof{figure}{\footnotesize{The first column shows the trajectories sampled from CVAE and VQ-VAE. These are then passed through the SF. The plots from second column onward show diverse feasible swarm coordination trajectories obtained after SF operations. The start and goal positions are marked with cross and red-dot respectively. }}
    \label{vq_cv_ours}
    
    \setcounter{figure}{1}

}{}{}
\makeatother

\begin{document} 

\title{\LARGE \bf Swarm-Gen: Fast Generation of Diverse Feasible Swarm Behaviors}

\author{Simon Idoko$^{1}$, B.Bhanu Teja$^{2}$, K.Madhava Krishna$^{2}$, Arun Kumar Singh$^{1}$
\thanks{$^{1}$: University of Tartu, Estonia. $2$: RRC, IIIT Hyderabad, India}
\thanks{Emails: simon.idoko@ut.ee, bbt1203@gmail.com, aks1812@gmail.com}
\thanks{Code-\href{https://github.com/cisimon7/SwarmGen}{https://github.com/cisimon7/SwarmGen}}
}

\maketitle

\thispagestyle{empty}
\pagestyle{empty}

\begin{abstract} 
Coordination behavior in robot swarms is inherently multi-modal in nature. That is, there are numerous ways in which a swarm of robots can avoid inter-agent collisions and reach their respective goals. However, the problem of generating diverse and feasible swarm behaviors in a scalable manner remains largely unaddressed. In this paper, we fill this gap by combining generative models with a safety-filter (SF). Specifically, we sample diverse trajectories from a learned generative model which is subsequently projected onto the feasible set using the SF. We experiment with two choices for generative models, namely: Conditional Variational Autoencoder (CVAE) and Vector-Quantized Variational Autoencoder (VQ-VAE). We highlight the trade-offs these two models provide in terms of computation time and trajectory diversity. We develop a custom solver for our SF and equip it with a neural network that predicts context-specific initialization. The initialization network is trained in a self-supervised manner, taking advantage of the differentiability of the SF solver.  We provide two sets of empirical results. First, we demonstrate that we can generate a large set of multi-modal, feasible trajectories, simulating diverse swarm behaviors, within a few tens of milliseconds. Second, we show that our initialization network provides faster convergence of our SF solver vis-a-vis other alternative heuristics.

\end{abstract}


\vspace{-0.1cm}
\section{Introduction}
\label{sec:introduction}
The ability to generate diverse swarm behaviors can have numerous applications. For example, it can be used to create a data-driven simulator to train navigation policies \cite{ren2019heter}, \cite{mavrogiannis2022b}, \cite{kazemkhani2024gpudrive}. It also helps to identify trajectories that meet multiple secondary criteria such as visibility \cite{lee2021target} and legibility \cite{capelli2019communication}.
In spite of its importance, there has been hardly any work on generating diverse multi-modal, feasible trajectories, especially in context of robot swarms. Existing distributed \cite{adajania2023amswarm}, \cite{adajania2024amswarmx}, \cite{soria2021distributed} or joint trajectory optimization approaches \cite{augugliaro2012generation}, \cite{rastgar2021gpu} are capable of generating only a single trajectory for each member of the swarm. Moreover, it is not straightforward to easily manipulate the trajectories these cited approaches generate.

In this paper, we leverage generative models to construct multi-modal swarm trajectories between a given start and goal state. We are inspired by the works on trajectory prediction that use generative models to capture the underlying distribution of the expert trajectories \cite{sriram2020smart}, \cite{ma2021continual}, \cite{xie2021congestion}, \cite{salzmann2020trajectron++}. During inference time, we can sample multiple trajectories from the learned model which subsequently leads to diverse swarm behaviors. However, what differentiates our approach from the likes of \cite{sriram2020smart}, \cite{ma2021continual}, \cite{xie2021congestion}, \cite{salzmann2020trajectron++}  is the focus on ensuring feasibility of the predicted trajectories, without sacrificing the ability to generate real-time swarm behaviors. The key innovations underlying our approach is described next.

\noindent \textbf{Algorithmic Contribution:}  We fit two generative models namely Conditional Variational Autoencoder (CVAE) and Vector-Quantized Variational Autoencoder (VQ-VAE) on the dataset of expert trajectories. Both these models provide different trade-offs in terms of computation time, trajectory diversity and success-rate. The trajectories sampled from learned CVAE or VQ-VAE may not perfectly satisfy the different kinematic and collision constraints. Thus, we propose a GPU accelerated, batched safety-filter(SF) that projects the sampled trajectories onto the the feasible set, in parallel. The SF forms the main computation bottleneck of our approach as it operates over the joint feasible space of all the robots. Thus, we learn a neural network that provides context-specific initialization for the SF to accelerate its convergence. We propose a self-supervised training set-up for the initialization network that leverages the differentiability of our SF solver. 


\noindent\textbf{Empirical Contribution:} We show that our approach can generate several multi-modal, feasible solutions that simulate diverse swarm coordination behaviors. We analyze the effectiveness of our generative approach with respect to the computation and memory budget. In many cases, we demonstrate real-time generative process on commodity GPUs. Finally, we demonstrate the effectiveness of our initialization policy in accelerating our custom SF solver vis-a-vis other simple hand-crafted warm-start choices. Although our prime focus in this paper is on swarm trajectories in 3D for drones, our method can be trivially extended for generating multi-modal behaviors in autonomous driving setting.


\section{Preliminaries}\label{sec:problem}
\subsubsection*{Notations} We will use normal font letters to represent scalars. The vectors and matrices will be represented by bold-faced lower and upper case respectively. We will use $\mathbf{p}_{{i|t}}$ to represent the 3D position of the $i^{th}$ robot at time-step $t$. The planning horizon will be represented by $H$.



\subsection{Constraint Description}
\noindent We consider the following set of constraints on the generated trajectories

\subsubsection*{Boundary Conditions} The generated trajectories should satisfy the following boundary constraints.
\begin{align}
    (\mathbf{p}_{i|0}, \dot{\mathbf{p}}_{i|0}, \ddot{\mathbf{p}}_{i|0})  = \mathbf{b}_{0}, \hspace{0.1cm} (\mathbf{p}_{i|H}, \ddot{\mathbf{p}}_{i|H}, \ddot{\mathbf{p}_{i|H}}) = \mathbf{b}_{H},\label{boundary_con}
\end{align}

\noindent where $\mathbf{b}_0$ and $\mathbf{b}_H$ are formed by stacking start and goal positions, velocities and accelerations respectively.

\subsubsection*{ Workspace Constraints} We require that the generated trajectories be contained within a spheroid shaped workspace. This can be enforced by the following quadratic constraints.
\begin{align}
    \lVert \mathbf{M}_{w}^{-1} \left(\mathbf{p}_{i|t} - \mathbf{p}_{w}\right)\rVert ^{2}_{2} - \mathbf{1} \leq \mathbf{0},\hspace{5mm} \forall t \hspace{3mm}  \forall i, \label{workspace_con}
\end{align}

\noindent where $\mathbf{p}_w$ is the center of the spheroid and $\mathbf{M}_{w}$ is a diagonal matrix formed with the axial dimension of the spheroid workspace $(a_w, a_w, b_w)$.

\subsubsection*{Inter-Robot Collision Avoidance} We model each robot as a spheroid. Thus, to generate safe swarm behaviors, we enforce the following non-convex quadratic constraints between each robot pair.
\begin{align}
    \lVert \mathbf{M}_{a}^{-1} \left(\mathbf{p}_{i, t} - \mathbf{p}_{j|t}\right)\rVert ^{2}_{2} - \mathbf{1} \geq \mathbf{0},\hspace{5mm}\forall k \hspace{3mm}  \forall i, j, \forall t   \label{inter_robot_con}
\end{align}
 \noindent where, $\mathbf{M}_{a}$ is a $3\times 3$ diagonal matrix formed with sum of individual robot axial dimensions $(\frac{a}{2}, \frac{a}{2}, \frac{b}{2})$.

\subsection{Polynomial Parametrization}
\noindent We parametrize the positional trajectory of the $i^{th}$ robot $\mathbf{p}_{i|0:H}$ in the following manner:

\begin{equation}
\begin{aligned}
    \mathbf{p}_{i| 0: H} = \left[\begin{array}{ccc}
        \mathbf{W} & \mathbf{0} & \mathbf{0} \\
        \mathbf{0} & \mathbf{W} & \mathbf{0} \\
        \mathbf{0} & \mathbf{0} & \mathbf{W}
    \end{array}\right] \left[\begin{array}{c}
        \mathbf{c}_{i, x} \\ \mathbf{c}_{i, y} \\ \mathbf{c}_{i, z} 
    \end{array}\right] = \overline{\mathbf{W}} \mathbf{c}_{i},
\end{aligned}
\label{eq::poly}
\end{equation}

\noindent where, $\mathbf{W}$ is a matrix formed with the time-dependent polynomial basis functions and the $\left[\mathbf{c}_{i,x}, \mathbf{c}_{i,y}, \mathbf{c}_{i,z}\right]$ is a vector of coefficients that define the trajectory. The velocities and accelerations can also be expressed in terms of the coefficients through the time dependent matrix $\dot{\overline{\mathbf{W}}}$, $\ddot{\overline{\mathbf{W}}}$


We roll the coefficients of all the robots together into a single vector $\boldsymbol{\xi} = (c_{1, x}, \dots, c_{n, x}, c_{1, y}, \dots, c_{n, y}, c_{1, z}, \dots, c_{n, z})$. Consequently, using \eqref{eq::poly}, we can re-write the constraints of the trajectory generation process in the following form
\begin{align}
    \mathbf{A}\, \boldsymbol{\xi} = \mathbf{b}, \qquad \mathbf{g}\left(\boldsymbol{\xi}\right) \leq \mathbf{0}
\end{align}


\noindent where the matrix $\mathbf{A}$ and vector $\mathbf{b}$ are constants. The first equality constraints is a compact representation of \eqref{boundary_con}. The function $\mathbf{g}$ is a vectorized re-phrasing of inequalities in \eqref{inter_robot_con}-\eqref{workspace_con}.


\section{Main Algorithmic Results}
\label{sec:methodology}

\begin{figure*}[h!]
    \centering
    \includegraphics[scale=0.4]{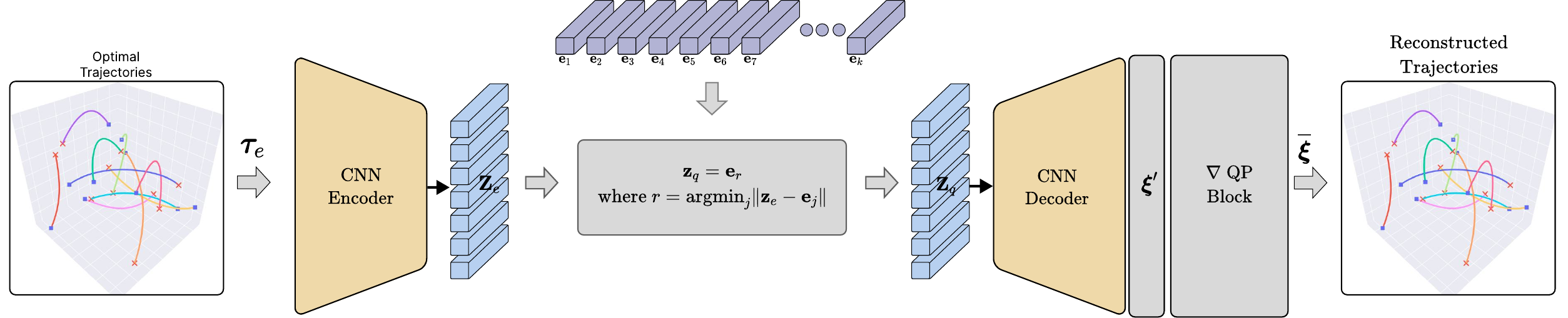}
    \caption{\footnotesize{Architecture of our VQ-VAE pipeline. A CNN encoder compresses optimal trajectories (time-stamped waypoints) into a discrete latent space $\mathbf{Z}_q$. The decoder reconstructs polynomial coefficients $\overline{\xi}$ based on $\mathbf{Z}_q$, which is then converted to trajectories. The differentiable QP block ensures that the reconstructed trajectories satisfy the boundary conditions on start and goal states. }}
    \label{fig_vqvae}
    \vspace{-0.2cm}
\end{figure*}

In this section, we present our core building blocks: the generative  models and the SF equipped with a initialization network.

\subsection{Learning VQ-VAE Prior Over Optimal Trajectories}
\noindent The first building block of our pipeline is a generative model that captures the underlying distribution of optimal trajectories. The basic approach is to compress the optimal trajectories to a latent space from which we can sample during the inference time. In this section, we use VQ-VAE to build this generative component because its discrete latent space is well suited to capture the multi-modality inherent in robots' trajectories.


Our VQ-VAE architecture is shown in Fig. \ref{fig_vqvae} and is a modified version of the original framework introduced in \cite{oord2017neural} for image generation. We employ a CNN encoder to compress optimal trajectories $\boldsymbol{\tau}_e$ into a continuous latent space $\mathbf{Z}_e \in \mathbb{R}^{L \times D}$, where $L$ represents the number of latent vectors, each with dimensionality $D$. Each latent vector(row of $\mathbf{Z}_e$ ), denoted as $\mathbf{z}_{e,i}$, is then transformed from continuous space to discrete space by mapping $\mathbf{Z}_e$ to $\mathbf{Z}_q$. This transformation is done using a latent embedding matrix $\mathbf{E} \in \mathbb{R}^{N \times D}$, also called the code-book, which consists of $N$ vectors $\mathbf{e}_j$. For each $\mathbf{z}_{e,i}$, the vector $\mathbf{e}_j$ is selected based on the nearest neighbor rule  \eqref{eqn_nearest}, creating the corresponding latent vector in $\mathbf{Z}_q$.
\begin{multline}
    \boldsymbol{z}_{q,i} = \boldsymbol{e}_r, \text{ where } r = \arg \min_j || \boldsymbol{z}_{e, i} - \boldsymbol{e}_j ||_2^2   \\
    \text{for } i \in \{1, 2, ..., L\} 
    \label{eqn_nearest}
\end{multline} 
The VQ-VAE decoder takes in $\textbf{Z}_q$ and transforms it into an intermediate vector $\boldsymbol{\xi}^'$. This is then fed to a differentiable QP layer to obtain the coefficient $\overline{\boldsymbol{\xi}}$ used to reconstruct the optimal trajectories. The role of the QP layer is to  ensure that the trajectories resulting from  $\overline{\boldsymbol{\xi}}$ satisfy the start and goal position, velocity and acceleration constraints. The structure of the QP is same as \eqref{eq::sf_form} but without the inequality constraints, and thus have a closed-form solution.

The VQ-VAE model is trained using a loss function defined in \eqref{eqn_vq_loss}, consisting of three components. The first component is the reconstruction loss which ensures that the encoder-decoder pair accurately reproduces the input trajectory. The other two components, the codebook loss (second term) and commitment loss(third term), serve to adjust the code-book vectors during training and handle the non-differentiable discretization process described in \eqref{eqn_nearest}.

\begin{multline}
    \mathcal{L}_{vqvae} = \| \mathbf{W} \boldsymbol{\xi} - \boldsymbol{\tau}_e \|_2^2 \;+\; \| \text{sg}[\mathbf{Z}_e] - \mathbf{E}\|_2^2 \\
    \;+\;  \beta \| \mathbf{Z}_e - \text{sg}[\mathbf{E}]\|_2^2
    \label{eqn_vq_loss}
\end{multline}

\subsubsection*{\textbf{PixelCNN Model}}
We modify the Conditional PixelCNN model \cite{van2016conditional} to enable sampling from the trained VQ-VAE. With this approach, we can generate diverse trajectories (via the VQ-VAE decoder) by conditioning the sampling process on features such as the start and goal positions, velocities, accelerations of the robots and workspace boundaries. We represent the conditioning vector as $\mathbf{s}$.

\begin{figure}[h!]
    \centering
    \includegraphics[height=3.7cm, width=8.6cm]{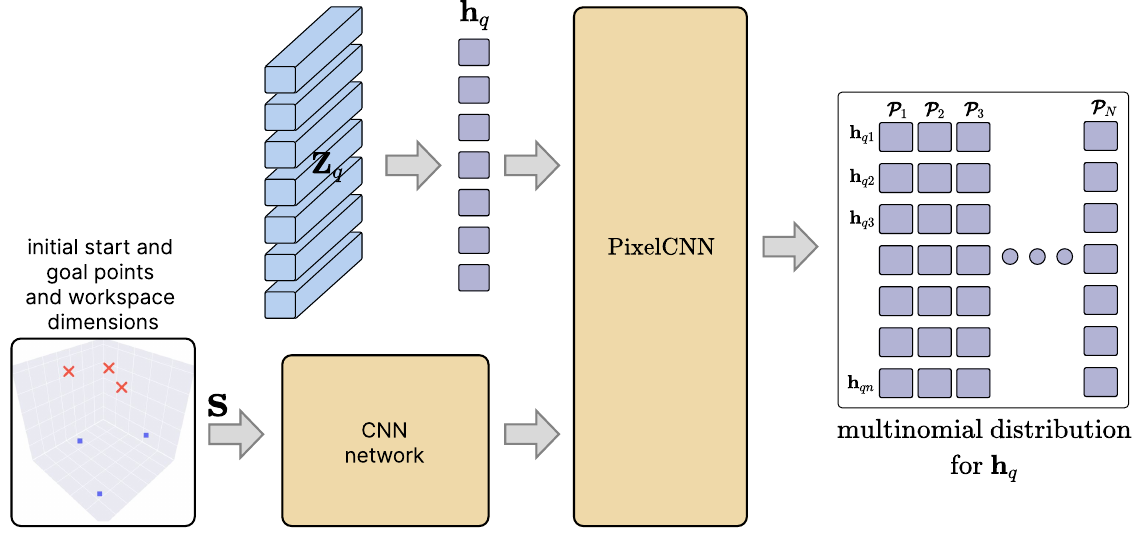}
    \caption{\footnotesize{ Architecture of PixelCNN used to sample from the VQ-VAE. Its output is a multinomial distribution over $\mathbf{h}_q$ which can be used to generate different samples of it. These are then used to generate different $\mathbf{Z}_q$ and consequently different multi-robot trajectories.    }}
    \label{pixel_cnn}
    \vspace{-0.2cm}
\end{figure}

To understand our PixelCNN model, recall that the discrete latent space, $\mathbf{Z}_q$, is structured as a matrix, where each row $\mathbf{z}_{q,i}$ is linked to the $r^{th}$ code-book vector using \eqref{eqn_nearest}. This makes it straightforward to compress the information from $\mathbf{Z}_q$ into a vector $\mathbf{h}_q$ , where each element stores the index of the code-book vector associated with its corresponding $\mathbf{z}_{q, i}$. The training process for the VQ-VAE described above gives us the ground truth values for $\mathbf{h}_q$. The PixelCNN model predicts a multinomial probability distribution over $\mathbf{h}_q$. For example, the first element of  $\mathbf{h}_q$ can be any integer between 0 to $N$ based on probability generated by PixelCNN (see Fig.\ref{pixel_cnn}).
Thus, sampling from this distribution allows us to create different versions of $\mathbf{h}_q$ which in turn, leads to different variations of $\mathbf{Z}_q$. These samples can then be passed through the VQ-VAE decoder, generating diverse trajectory outputs.

A key characteristic of the PixelCNN model is that it generates the probability distribution over $\mathbf{h}_q$ in an auto-regressive fashion, using \eqref{auto_regressive}. This means that the prediction for each element in $\mathbf{h}_q$ is influenced by the predictions of the preceding elements, as well as the conditioning vector. During training, a cross-entropy loss is computed between the true values of $\mathbf{h}_q$ (obtained from VQ-VAE) and the values predicted by PixelCNN. This loss is used to optimize the model's parameters for the multinomial distribution $\mathcal{P}$.
\vspace{-0.3mm}
\begin{align}
 \mathcal{P}(\mathbf{h}_q|\mathbf{s}) = \prod_{i=1}^{L}p(\mathbf{h}_{q,i}|\mathbf{h}_{q,1},\hdots, \mathbf{h}_{q,i-1}, \mathbf{s})
    \label{auto_regressive}
\end{align}


\begin{figure*}
    \centering
    \includegraphics[scale = 0.22]{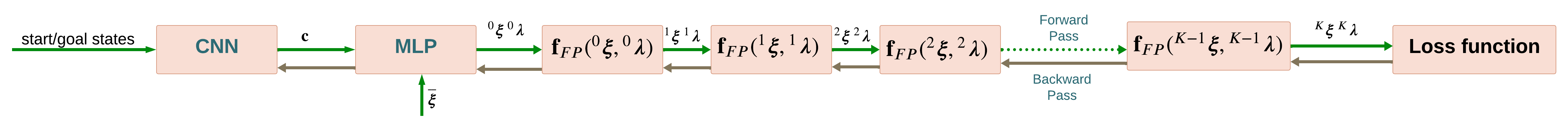}
    \caption{Training pipeline for learning context specific initialization for our SF solver. It consists of learnable layers (CNN+MLP) followed by an unrolled chain of our fixed-point solver underlying our SF. During training, the gradients of the loss function is traced through the fixed-point solver. This in turn ensures that the CNN and MLP layers are aware of how its predictions are leveraged by the downstream fixed-point solver.}
    \label{fig:learned_SF}
    \vspace{-0.5cm}
\end{figure*}

\subsection{CVAE Prior Over Optimal Trajectories}
\noindent The auto-regressive nature of PixelCNN implies that sampling from a trained VQ-VAE would be computationally expensive, especially when the number of code-book vectors is large. Thus, in this sub-section, we present a CVAE based model, as a cheaper alternative. The latent space of CVAE is  Gaussian and thus is amenable to faster sampling . However, on the other hand, the Gaussian latent space may not fully capture the multi-modality of the optimal trajectories.

\begin{figure}[h!]
    \centering
    \includegraphics[scale=0.4]{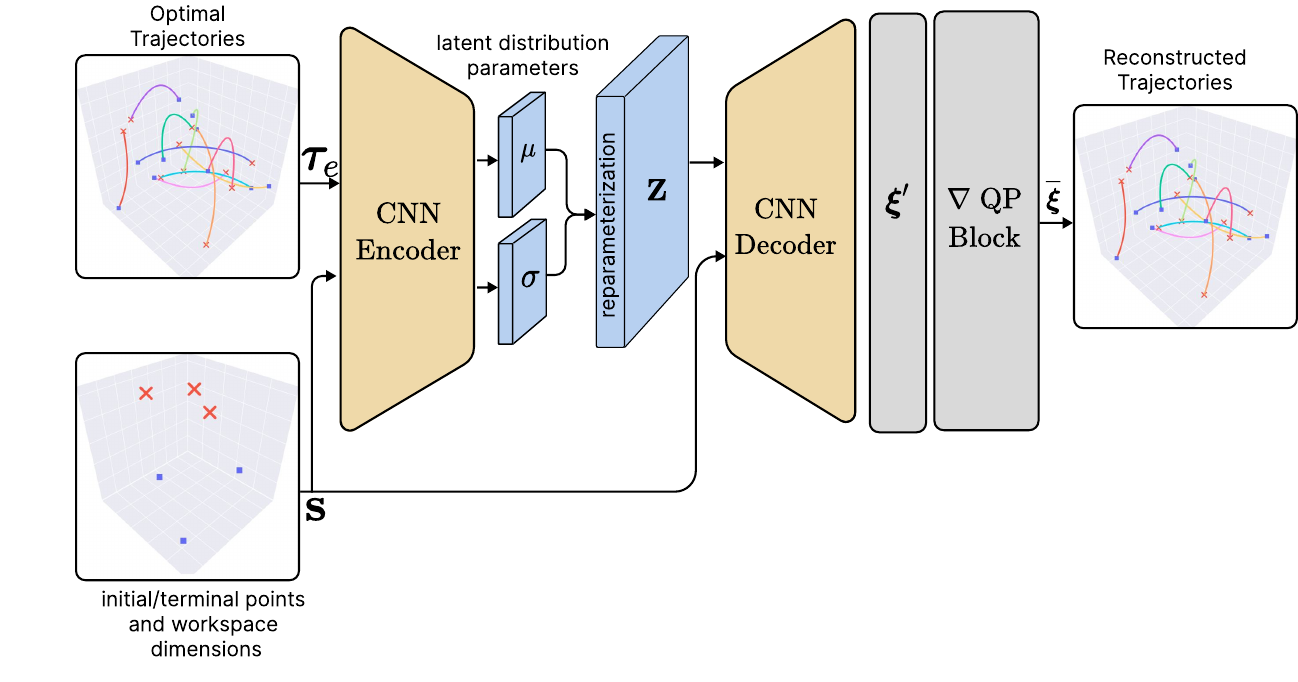}
    \caption{\footnotesize{Our pipeline to fit CVAE to the dataset of expert multi-robot trajectories. The differentiable QP block is simply \eqref{eq::sf_form} without the inequality constraints. }}
    \label{fig_cvae}
    \vspace{-0.2cm}
\end{figure}

Our CVAE architecture, illustrated in Fig.\ref{fig_cvae} and it builds upon the model introduced in \cite{sohn2015learning}. The core idea in CVAE is to compress the expert trajectory into a Gaussian latent space. Thus, in our pipeline, the expert trajectory $\boldsymbol{\tau}_e$ and the conditioning variable $\mathbf{s}$, consisting of start and goal state are fed to a CNN based encoder that  outputs the mean ($\boldsymbol{\mu}$) and standard deviation ($\boldsymbol{\sigma}$) of the latent variable $\mathbf{z}$.



The decoder is again based on CNN. It receives the latent variable $\boldsymbol{Z}$ and the conditioning variable $\mathbf{s}$ as inputs and produces an intermediate output for the trajectory coefficients ${\boldsymbol{\xi}}^'$ that is passed to a differentiable QP layer. The output of the QP layer is $\overline{\boldsymbol{\xi}}$ which is used to reconstruct back the trajectories. We recall that similar to VQ-VAE setting, the differentiable QP layer simply modifies  ${\boldsymbol{\xi}}^'$ to ensure that the trajectories resulting from $\overline{\boldsymbol{\xi}}$ satisfies the start and goal position, velocity and acceleration constraints.

Both the encoder and decoder are trained in an end-to-end fashion with the  training loss defined in \ref{eqn_cv_loss}. The first term in \eqref{eqn_cv_loss} is the typical mean-squared-error loss that promotes faithful reconstruction. The second-term is the Kullback Liebler (KL) divergence loss that regulates the latent variable $\mathbf{z}$ to be as close as possible to a unit-normal Gaussian.

\begin{align}
    \mathcal{L}_{cvae} = \| \mathbf{W} \boldsymbol{\xi} - \boldsymbol{\tau}_e \|_2^2 \;+\; KL[q_{\phi}(\mathbf{Z}|\boldsymbol{\tau},\mathbf{s}), | \mathcal{N}(0|I)]
    \label{eqn_cv_loss}
\end{align} 

\subsection{Safety-Filter(SF)}
\noindent Our SF is defined by the following optimization problem.

\begin{equation}
\begin{aligned}
    \min_{{\boldsymbol{\xi}}} \frac{1}{2}\left \Vert {\boldsymbol{\xi}}-\overline{\boldsymbol{\xi}}\right\Vert_2^2\\ 
    \hspace{5mm} \mathbf{A}\, {\boldsymbol{\xi}} = \mathbf{b} , \qquad \mathbf{g}\left({\boldsymbol{\xi}}\right) \leq \mathbf{0}   
\end{aligned}
\label{eq::sf_form}
\end{equation}

\noindent The proposed SF computes the minimal correction required to be made to VQ-VAE/CVAE predicted trajectory coefficient $\overline{\boldsymbol{\xi}}$  to satisfy the collision and workspace constraints. We propose a custom SF solver that reduces the solution process of \eqref{eq::sf_form} into a fixed-point operation of the following form, where the left subscript $k$ denotes the iteration number.
\begin{align}
    {^{k+1}}{\boldsymbol{\xi}}, {^{k+1}}\boldsymbol{\lambda} = \mathbf{f}_{FP}({^{k}}{\boldsymbol{\xi}}, {^{k+1}}\boldsymbol{\lambda})
    \label{fixed_point}
\end{align}

\noindent We derive the mathematical structure of $\mathbf{f}_{FP}$ in the Appendix \ref{appendix}, where we also define the Lagrange multiplier $\boldsymbol{\lambda}$. But some interesting points are worth pointing out immediately. First, by carefully reformulating the quadratic constraints \eqref{inter_robot_con}-\eqref{workspace_con}, we can ensure that the numerical computations underlying $\mathbf{f}_{FP}$ only requires matrix-matrix products, which can be easily batched and accelerated over GPUs. Secondly, $\mathbf{f}_{FP}$ involves only differentiable operation, which allows us to compute how changes in the initialization of $\mathbf{f}_{FP}$ affect the convergence process. This differentiability forms the core of our learning pipeline designed to accelerate the convergence of fixed-point iteration \eqref{fixed_point}. 


\subsubsection*{\textbf{Learned Initialization for SF}} To train an initialization network for the SF solver in a self-supervised manner, we consider a hybrid architecture  shown in Fig.\ref{fig:learned_SF}. It consists of a learnable part followed by an unrolled chain of length $K$ of fixed-point iteration. The CNN layer takes the information from the start and goal positions, velocities and accelerations and produces an encoding $\mathbf{c}$. This is then stacked together with the trajectory coefficient $\overline{\boldsymbol{\xi}}$ predicted by CVAE/VQ-VAE to form the complete context vector $\mathbf{v}$. An MLP, then processes the context vector to get the initialization $^{0}{\boldsymbol{\xi}}, ^{0}\boldsymbol{\lambda}$. Let $^{K}{\boldsymbol{\xi}}, ^{K}\boldsymbol{\lambda}$ be the solution obtained by running the fixed-point iteration for $K$ iterations from the predicted initialization. We formulate the following optimization problem to train the learnable part of the SF. 
\begin{align}
    \min_{\boldsymbol{\Theta}} \sum_{k=0}^{k-1} \left \Vert \begin{bmatrix}
        {^{k+1}}{\boldsymbol{\xi}}\\
        {^{k+1}}\boldsymbol{\lambda}
    \end{bmatrix}-\mathbf{f}_{FP} ({^{k}}{\boldsymbol{\xi}}, {^{k}}\boldsymbol{\lambda}) \right\Vert_2^2+\left\Vert {^K}{\boldsymbol{\xi}}-\overline{\boldsymbol{\xi}}\right\Vert_2^2,
    \label{NN_loss}
\end{align}

\noindent where $\boldsymbol{\theta}$ contains the weights of the CNN and MLP modules. The first term in \eqref{NN_loss} minimizes the fixed-point residual at each iteration and is responsible for accelerating the convergence of \eqref{fixed_point}  \cite{sambharya2024learning}. The second term ensures that the initialization predicted from the MLP leads to a solution that is minimally displaced from the original VQ-VAE/CVAE predicted value. During training, the gradient of the loss function is traced through the stacked layers of $\mathbf{f}_{FP}$ to the learnable CNN and MLP layers. This ensures that the neural network layers are aware of how its predictions are leveraged by the downstream solver and leads to highly effective warm-start for the fixed-point solver. 





\section{Connection To Existing Works}\label{rel_work}
\noindent \textbf{Multi-Agent Trajectory Prediction:} Our approach is related to the literature on predicting multi-agent trajectories in autonomous driving \cite{sriram2020smart}, \cite{ma2021continual}, \cite{xie2021congestion}, \cite{salzmann2020trajectron++} and crowd/pedestrian behavior modeling \cite{zhao2019multi}, \cite{yuan2021agentformer}. The core of many of these works are indeed generative models. For example, \cite{sriram2020smart}, \cite{salzmann2020trajectron++}, \cite{yuan2021agentformer} leverage CVAE to predict multiple trajectories. Unfortunately, translating these cited works for generating diverse robot swarm behaviors has many challenges. For example, since these works focus on trajectory predictions, they are not designed for generating multiple swarm behavior between the same start and goal positions. Moreover, they do not have explicit mechanism to ensure that the predicted trajectories satisfy workspace and collision avoidance constraints. 

Our approach is closest to a recent diffusion based model presented in \cite{jiang2023motiondiffuser}. The inference-time sampling from diffusion models can be guided through additional cost terms modeling goal reaching and collision avoidance. However, tuning the effect of each cost term could be tricky. Although,  it is theoretically possible to extend \cite{jiang2023motiondiffuser} for predicting multi-modal swarm behaviors,  
our work provides a better alternative in the following ways. First, our SF based constraint satisfaction requires no tuning and can be accelerated using learning. Second and more importantly, diffusion models can be notoriously slow, especially while sampling a large number of samples from them. In contrast, our approach can be used online. Moreover, we show that due to our SF, simple generative models like CVAE can provide exceptionally effective while being real-time on commodity GPUs.

\noindent \textbf{Learning to Warm-start Optimization/FP Iteration:} The simplest approach towards learning good initialization for optimization/FP solvers is to fit some neural network model over the  dataset of optimal solution. Subsequently, the predictions from the learned model can be used to initialize the same optimizer/FP solver from where the dataset was generated \cite{celestini2024transformer}, \cite{pulver2021pilot}. For example, in our context, this would entail storing the solution of \eqref{fixed_point}, fitting a model to it and subsequently leveraging its predictions for warmstart. However, such approaches often fall short in practice. This is because, the notion of "good initialization" is very solver-specific \cite{sambharya2024learning}. In other words, initializations that work for an interior-point solver may not provide any computational gains if we replace it with a gradient descent routine. Thus, authors in \cite{sambharya2024learning} recommend hybrid architectures like that shown in Fig.\ref{fig:learned_SF} wherein, during the training process, the neural network layers are aware of how its predictions are used by the downstream solver. Our work extends \cite{sambharya2024learning} to the non-convex multi-robot trajectory optimization setting. 

\noindent \textbf{Contribution Over Author's Prior Works:} The proposed work extends \cite{shrestha2023end}, \cite{idoko2024learning} to the multi-robot setting. Moreover, our SF solver is an improved and batched version of the optimizer presented in \cite{rastgar2021gpu}.

\section{Validation and Benchmarking}\label{val}
The objective of this section is twofold. First, we demonstrate that our approach is indeed capable of multi-modal feasible swarm trajectories in a scalable manner. In this context, we study trade-offs between our CVAE and VQ-VAE based approach. Second, to show the role of the initialization network in accelerating the convergence of SF solver.

\subsection{Implementation Details}
\noindent The VQ-VAE, CVAE, PixelCNN and the SF were all trained using Pytorch. However, for faster inferencing, the learned SF was converted to JAX. For data collection, we uniformly sample start and goal positions from different workspaces of varying dimensions centered around the origin. The start and goal velocities and accelerations wee always kept at zero. We used an improved version of \cite{rastgar2021gpu} to generate optimal trajectories between the sampled start and goal pairs. These were used to train VQ-VAE and CVAE. 

\subsubsection*{VQ-VAE Network Details} 
The VQ-VAE encoder consists of 5 layers, each with a convolution layer (128 output channels), ReLU activation, and batch normalization. The model has 512 codebook vectors ($N$), each with a 3-dimensional vector ($D$). The latent vector size ($L$) is 25 for the 4- and 8-agent models, and 100 for the 16-agent model to accommodate higher complexity. The VQ-VAE decoder has 4 convolution transpose layers, each with 128 output channels, ReLU activation, batch normalization, and dropout layers.

\subsubsection*{CVAE Network Details}
The CVAE encoder, like the VQ-VAE, has 5 layers with a convolution layer (128 output channels), ReLU activation, and batch normalization. The latent vector ($\mathbf{Z}$) size is 25 for the 4- and 8-agent models, and 100 for the 16-agent model, with each vector having 3 dimensions. This is flattened and passed through two MLP layers to compute the mean and standard deviation of the latent vector distribution. The CVAE decoder consists of 4 convolution transpose layers (128 output channels), leaky ReLU, batch normalization, and dropout layers. Additionally, a network with two convolution layers (ReLU activation and batch normalization) processes the state vector to be concatenated with the latent space vector before being passed to the decoder.

\subsubsection*{Initialization Network Details}
The CNN layer of the initialization network (recall Fig.\ref{fig:learned_SF}) is similar to PointNet architecture but as 18 input channels. The start and goal position, velocity and accelerations in 3D forms an 18 dimensional vector. These are stacked for all the agents and passed through the input channels of the CNN to obtain a feature vector $\mathbf{c}$. This is then concatenated with the VQ-VAE/CVAE outputs and passed through a 4-layer MLP consisting of linear layers, LeakyReLU activation, and batch normalization. More model details are shown in the accompanying video.

\subsection{Generating Multi-Modal Swarm Trajectories: CVAE Vs VQ-VAE}
\noindent Fig.\ref{vq_cv_ours} shows a typical result obtained with our approach. The first step is to sample a batch coefficients from the learned CVAE/VQ-VAE which are then converted to trajectories using \eqref{eq::poly} (first column of Fig.\ref{vq_cv_ours}). We then pass these coefficients through the SF optimizer \eqref{eq::sf_form} that projects all the sampled coefficients onto the feasible set, in parallel. The output of the SF are typically polynomial coefficients that define multi-modal, feasible trajectories. A few samples of the diverse swarm behavior is shown in Fig.\ref{vq_cv_ours}(b)-(g), (g)-(j). For example, consider the robot with magenta trajectory shown in Fig\ref{vq_cv_ours}(g),(j). In Fig.\ref{vq_cv_ours}(g), the robot flies over the other neighbors to avoid obstacles, while Fig.\ref{vq_cv_ours}(h), it files closer to the ground.

\begin{figure}[h!]
    \centering
    \includegraphics[scale = 0.30]{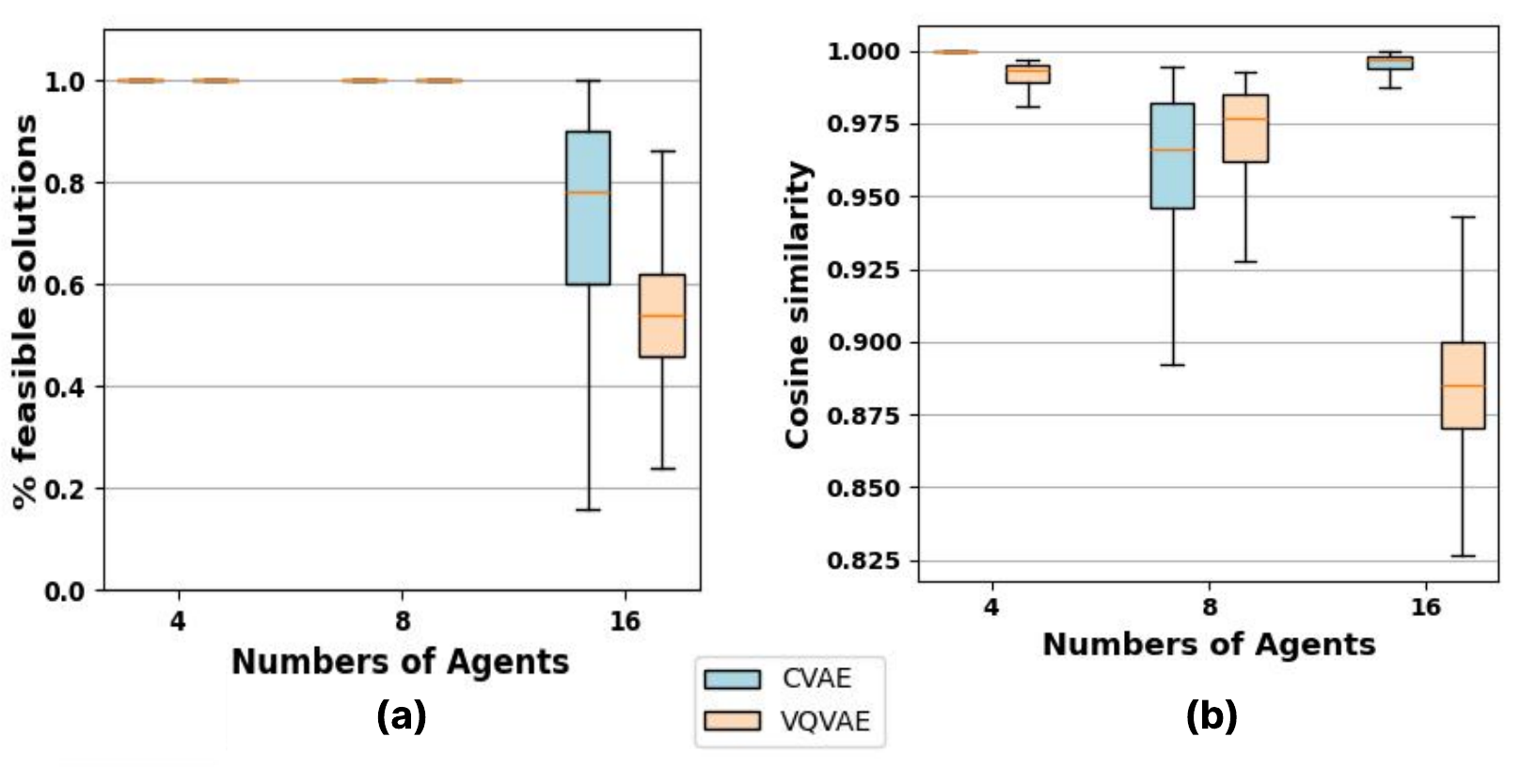}
    \caption{\footnotesize{Fig.(a) represents the fraction of the trajectories sampled from VQ-VAE/CVAE that becomes feasible after applying SF for 200 iterations. Fig.(b) compares the diversity of CVAE and VQ-VAE models through the cosine similarity metric computed over the feasible trajectories.   }}
    \label{fig_mod_theshold}
    \vspace{-0.2cm}
\end{figure}

\noindent \textbf{Diversity:} To further compare the generated behaviors with CVAE and VQ-VAE, we generated 3000 random start and goal positions for different number of robots. For each pair, we sampled 50 trajectories from the learned CVAE and VQ-VAE. We then apply SF for 200 iterations to improve the constraint satisfaction of the generated trajectories. Fig.\ref{fig_mod_theshold} (a) expresses the number of feasible solutions obtained after SF operations, as a fraction of total number of sampled trajectories (50). As can be seen, in all the cases, we obtained multiple feasible solutions. Moreover, CVAE produced more feasible solutions. However, more solutions doe not necessarily translate to more diversity. As shown in Fig.\ref{fig_mod_theshold}(b), the cosine similarity of VQ-VAE based model is lower than that based on CVAE, implying that VQ-VAE leads to a more diverse swarm behavior. This in turn, can be directly attributed to the discrete latent space of VQ-VAE that is more capable of capturing multi-modality.


\begin{figure}[h!]
    \centering
    \includegraphics[scale = 0.21]{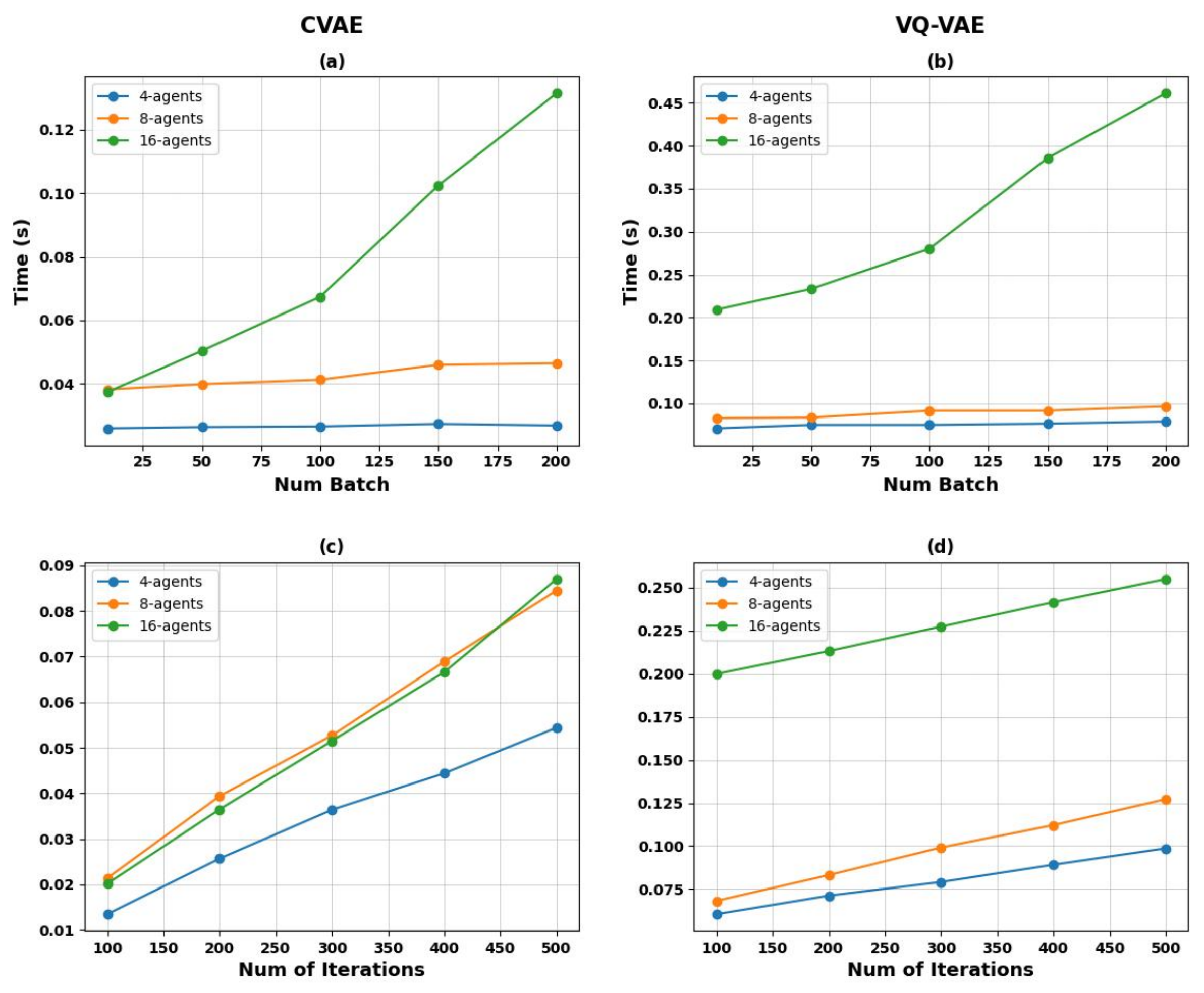}
    \caption{\footnotesize{Fig(a)-(b): Computation time with respect to batch size (number of sampled trajectories from VQ-VAE/CVAE) when the SF is ran for 200 iterations. Fig.(c)-(d): Computation time with respect to the number of SF iterations for a given batch size of 10. }}
    \label{fig_comp_time} 
    \vspace{-0.2cm}
\end{figure}

\noindent \textbf{Computation Time:} Fig.\ref{fig_comp_time} compares the computation time of CVAE and VQ-VAE based approach with respect to the number of iterations of SF and batch size(number of sampled trajectories). All the timings were obtained on a RTX 3090 GPU. For smaller batch size ($\approx 10$), both CVAE and VQ-VAE approaches can be run online, with the former capable of running at interactive-rates. For up to 8 agents, the VQ-VAE based approach is only marginally slower than the CVAE counterpart. The larger computation time required for 16 agents can be attributed to the fact that it required a VQ-VAE with a more complex latent space. At larger batch size of 200, VQ-VAE based swarm behavior generation can be done at approximately 2Hz, which is around 10 times slower than that achieved by CVAE based approach. Interestingly, the computation time of both VQ-VAE and CVAE show a very scalable increase with respect to the batch size. This is due to excellent vectorized and GPU accelerated inferencing of the learned neural network models as well as the computations underlying the SF.  

Fig.\ref{fig_comp_time}(c)-(d) show the variation of the computation time with respect to number of SF iterations for a batch size of 10. As can be seen, for both CVAE and VQ-VAE based approach, we get an almost linear growth in run-time. This trend is important as sometimes we need to run the SF for more iterations to ensure feasibility of the generated trajectories. However, in all our experiments, running SF for around 200 iterations proved enough to get a large number of feasible solutions in any given scene (recall Fig.\ref{fig_mod_theshold}(a)).

\noindent \textbf{Summarizing CVAE and VQ-VAE trends}
\begin{itemize}
    \item VQ-VAE based pipeline can generate more diversity in swarm behaviors. For up to 8 agents, it can be run at an interactive rates, even with a large batch size. 
    \item For larger number of agents, VQ-VAE can be still be preferred choice when generating a small number of swarm behaviors(around 10).
    \item When generating a large number of scenarios for a large swarm size, CVAE can be a good alternative, especially if there are hard run-time constraints. \item We can create an ensemble of CVAE and VQ-VAE based predictions. 
\end{itemize}

\subsection{Validating the Efficacy of SF Initialization Network  } 
\begin{figure}[h!]
    \centering
    \includegraphics[scale = 0.21]{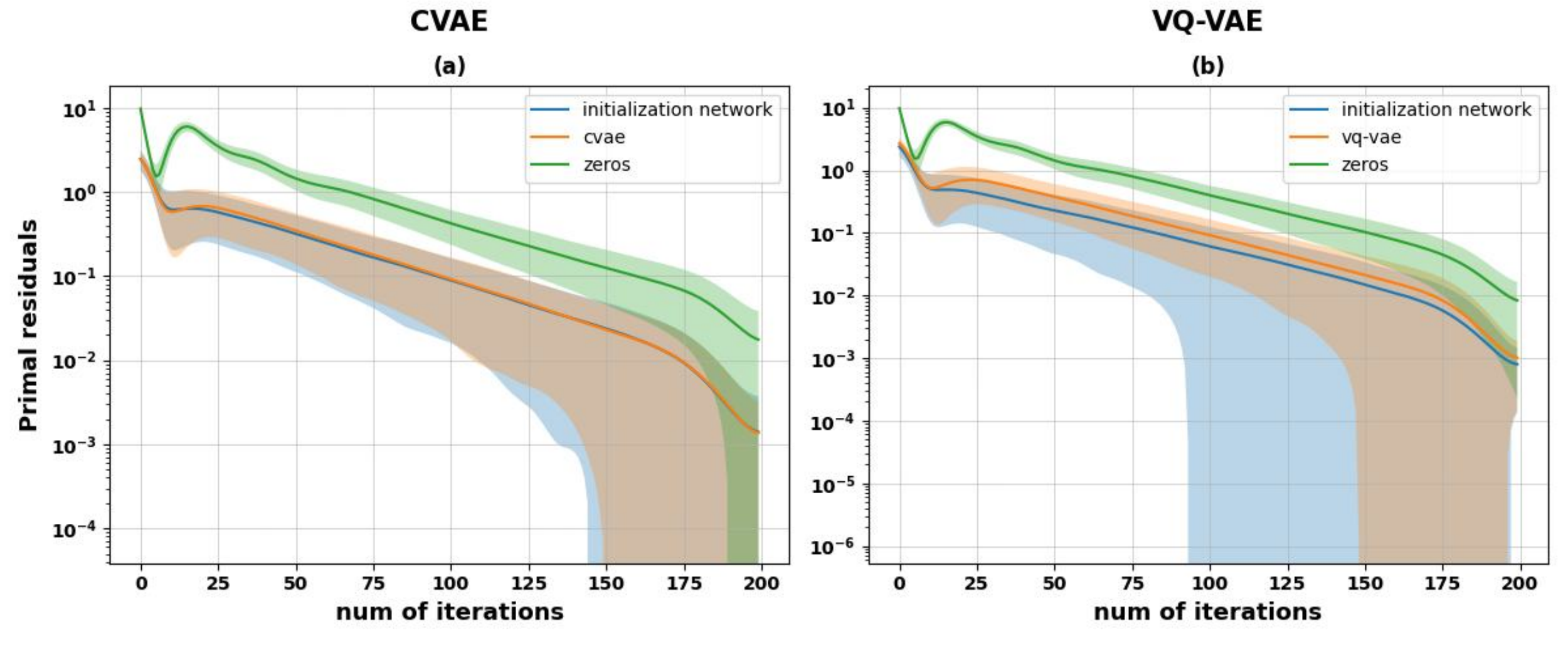}
    \caption{\footnotesize{Primal residuals of the SF solver with different initialization strategies.}}
    \label{fig_res}
    \vspace{-0.8cm}
\end{figure}

\noindent Fig.\ref{fig_res} shows the primal residuals of SF solver across iterations for different initialization strategies. Its mathematical formula is defined in  \eqref{primal_residual}(Appendix) and dictates the feasibility of the trajectories. As can be seen, the fastest reduction in the primal residuals is observed when the SF solver is initialized with the network trained using \eqref{NN_loss}. In comparison, the residual trends obtained when initialized with a naive zero vector is demonstrably worse. When the SF is directly initialized with CVAE/VQ-VAE samples, the residuals shows a mixed trend. The VQ-VAE samples are poor initializers for SF, while CVAE samples show competitive performance. This unreliability is not surprising as neither VQ-VAE nor CVAE training is aware of how SF operates. To obtain consistently good warm-start performance, it is important to embed the SF fixed-point solver within the learning pipeline (Fig.\ref{fig:learned_SF}), which ensures that the neural network layers are aware of how its predictions are leveraged by the downstream solver during training. 



\section{Conclusion}\label{sec:conclusion}
We tackled an important but relatively unaddressed problem of generating multiple feasible and diverse trajectories for robot swarms. At an abstract level, this is similar to identifying diverse solutions of a highly non-convex problem. Our approach was based on combining generative models with a SF based inference-time corrections. Moreover, to ensure computational efficiency, we also trained an initialization network that dramatically accelerated the convergence of our custom SF solver. We empirically analyzed the use of CVAE and VQ-VAE as generative models from the point of view of trajectory diversity and computation time. We believe our approach is a first of its kind that opens up new possibilities in data-driven simulation and multi-robot coordination. Although not presented here, we also did extensive experimentation with diffusion models. But the slow diffusion sampling, especially for larger batch size implied slow generative process. We believe VQ-VAE in conjunction with SF provides a good trade-off between model complexity and trajectory diversity. In future, we aim to apply our approach for controlled traffic generation for autonomous driving simulation and develop ways to condition the generative process on natural language descriptions. For the former application, we are looking to bring in ways to enforce road-structure or map constraints into our generative process. 


\section{Appendix}\label{appendix} 
In this section, we present a detailed derivation of our SF optimizer. We extend the derivation of \cite{rastgar2021gpu} to incorporate workspace constraints (\eqref{workspace_con}) and show that our SF optimizer steps are differentiable and can be run in a batched fashion over GPUs.

\subsection{Custom SF optimizer}

\subsubsection{Quadratic Constraints in Spherical Form} The inter-robot collision avoidance constraints \eqref{inter_robot_con} can be re-phrased into the following spherical form.

\small
\begin{dmath}
    \mathbf{p}_{i|k}-\mathbf{p}_{j|t} = d_{ij|t}\begin{bmatrix}
        a \cos\alpha_{ij|t}\sin\beta_{ij|k}\\  a \sin\alpha_{ij|t}\sin\beta_{ij|t} \\ b \cos\beta_{ij|t}
    \end{bmatrix}, 1\leq d_{ij|t}\leq \infty,
    \label{inter_agent_polar}
\end{dmath}
\normalsize
\noindent where $\alpha_{ij|t}, \beta_{ij|t}$ and $ d_{ij|t}$ are the spherical angles and normalized line of sight-distance between the agents. These are unknown and will be obtained by the SF optimizer along with other variables.

Following a similar notation, we can re-write the workspace constraints \eqref{workspace_con} in the following manner.

\small
\begin{dmath}
    \mathbf{p}_{i|t}-\mathbf{p}_{w} = d_{w,i|t}\begin{bmatrix}
        a_w \cos\alpha_{w,i|t}\sin\beta_{w,i|t}\\  a_w \sin\alpha_{w,i|t}\sin\beta_{w,i|t} \\ b_w \cos\beta_{w,i|t}
    \end{bmatrix}, 0\leq d_{w,i|t}\leq 1,
    \label{workspace_polar}
\end{dmath}
\normalsize
\subsubsection{Reformulated Problem}
\noindent Using \eqref{inter_agent_polar}-\eqref{workspace_polar} and the polynomial representation \eqref{eq::poly}, we get the following reformulation of the SF optimizer.
\small
\begin{align}
    \min_{\boldsymbol{\xi}, \Tilde{\boldsymbol{\alpha}}, \Tilde{\mathbf{d}}, \Tilde{\boldsymbol{\beta}}} \frac{1}{2}\left \Vert \boldsymbol{\xi}-\overline{\boldsymbol{\xi}}\right\Vert_2^2 \label{reform_cost}\\
    \mathbf{A}\boldsymbol{\xi} = \mathbf{b} \label{reform_eq}\\
    \mathbf{F}\boldsymbol{\xi} = \mathbf{e}(\Tilde{\boldsymbol{\alpha}},\Tilde{\boldsymbol{\beta}}, \Tilde{\mathbf{d}})\label{reform_non_convex}  \\
    \mathbf{d}_{min}\leq \Tilde{\mathbf{d}} \leq \mathbf{d}_{max}
\end{align}
\normalsize
 \small
  \begin{align}
      {\textbf{F}}\hspace{-0.10cm} =\hspace{-0.15cm} \begin{bmatrix}
        \begin{bmatrix}
        \mathbf{F}_{o} \\
        \mathbf{F}_w 
    \end{bmatrix} \hspace{-0.25cm}&\hspace{-0.25cm} \textbf{0} \hspace{-0.25cm}&\hspace{-0.25cm} \textbf{0} \\ \textbf{0}\hspace{-0.25cm} & \hspace{-0.25cm}\hspace{-0.15cm}\begin{bmatrix}
         \mathbf{F}_{o} \\
        \mathbf{F}_w 
    \end{bmatrix} \hspace{-0.25cm}& \hspace{-0.25cm} \textbf{0}\\
         \textbf{0}\hspace{-0.25cm}& \hspace{-0.25cm}\textbf{0} \hspace{-0.25cm}&\hspace{-0.25cm} \begin{bmatrix}
         \mathbf{F}_{o} \\
        \mathbf{F}_w 
    \end{bmatrix}
    \end{bmatrix}\hspace{-0.15cm}, \hspace{-0.05cm}
    {\mathbf{e}} \hspace{-0.1cm}= \hspace{-0.15cm}\begin{bmatrix}
      a\mathbf{d}\cos{ \boldsymbol{\alpha}} \sin{ \boldsymbol{\beta}} \\ 
      x_w+a_w\mathbf{d}\cos{ \boldsymbol{\alpha}_w} \sin{ \boldsymbol{\beta}_w} \\ 
      a\mathbf{d}\sin{ \boldsymbol{\alpha}} \sin{ \boldsymbol{\beta}} \\
      y_w+a_w\mathbf{d}\sin{ \boldsymbol{\alpha}_w} \sin{ \boldsymbol{\beta}_w}\\
      b\hspace{0.1cm} \mathbf{d}\cos{ \boldsymbol{\beta}} \\
      z_w+b_w\mathbf{d}_w\cos{ \boldsymbol{\beta}_w}    
      \end{bmatrix}\hspace{-0.15cm},\label{f_e}
  \end{align} 
  \normalsize

\noindent $\Tilde{\boldsymbol{\alpha}} = (\boldsymbol{\alpha}, \boldsymbol{\alpha}_w)$, $\Tilde{\boldsymbol{\beta}} = (\boldsymbol{\beta}, \boldsymbol{\beta}_w)$, $\Tilde{\mathbf{d}} = (\mathbf{d}, \mathbf{d}_w)$. The $\boldsymbol{\alpha}$ is formed by stacking $\alpha_{ij|t}$ for all agent pairs $(i, j)$ and all time step $t$. Similarly, $\boldsymbol{\alpha}_w$ is formed by stacking $\alpha_{w, i|t}$ for all $i$ and $t$. We follow similar construction for  $\boldsymbol{\beta} $, $\boldsymbol{\beta}_w$, $\mathbf{d}$, and $\mathbf{d}_w$. The constants $(x_w, y_w, z_w)$ are the components of the workspace center $\mathbf{p}_w$. The matrix $\mathbf{F}$ is defined in the following manner.

\small
\begin{align}
    \mathbf{F}_o = \begin{bmatrix}
    \mathbf{F}_{o, 1} \\
    \mathbf{F}_{o, 2}\\
    \vdots\\
    \mathbf{F}_{o, n-1}
  \end{bmatrix}\otimes \mathbf{P}, \qquad \mathbf{F}_w = \begin{bmatrix}
    \mathbf{P} & & \\
    & \ddots & \\
    & & \mathbf{P}
  \end{bmatrix}
\end{align}
\normalsize

\small
\begin{align}
    \mathbf{F}_{o, i} = \begin{bmatrix}
        \mathbf{F}_i & -\mathbf{I}
    \end{bmatrix}, \mathbf{F}_i = \begin{bmatrix}
        \mathbf{0}_{n-i\times 1} & \mathbf{0}_{n-i\times 1} &\dots &{\mathbf{1}_{n-i\times 1}}
    \end{bmatrix}_{n-i\times i}
\end{align}
\normalsize

\noindent The matrix $\mathbf{F}_w$ is a block-diagonal with number of blocks equal to the number of agents. The matrix $\mathbf{F}_i$ is formed with $i-1$ blocks of $n-i$ length zero vector and a single vector of ones at the $i^{th}$ column. The symbol $\otimes$ represents the Kronecker product. 

\subsubsection{Solution Process} We relax the non-convex equality constraints \eqref{reform_non_convex} as penalties and augment them into the cost function using the Augmented Lagrangian method 
\begin{align}
    \mathcal{L} = \frac{1}{2}\left \Vert \boldsymbol{\xi}-\overline{\boldsymbol{\xi}}\right\Vert_2^2+\frac{\rho}{2}\left \Vert \mathbf{F}\boldsymbol{\xi}-\mathbf{e}(\Tilde{\boldsymbol{\alpha}}, \Tilde{\boldsymbol{\beta}}, \Tilde{\mathbf{d}} )\right\Vert_2^2-\langle \boldsymbol{\lambda}, \boldsymbol{\xi}\rangle,
    \label{aug_lag}
\end{align}
\noindent where $\rho$ is a known constant and the variable $\boldsymbol{\lambda}$ are so-called Lagrange multipliers. We minimize \eqref{aug_lag} subject to \eqref{reform_eq} through an Alternating Minimization (AM) approach, wherein at each step, only one variable group among $\boldsymbol{\xi}, \Tilde{\boldsymbol{\alpha}}, \Tilde{\boldsymbol{\beta}}, \Tilde{\mathbf{d}}$ is optimized while others are held fixed. Specifically, the AM routine decomposes into the following iterative steps, wherein the left superscript $k$ tracks the values of a variable across iterations. For example, ${^k}\boldsymbol{\xi}$ is the value of $\boldsymbol{\xi}$ at iteration $k$.

\small
\begin{subequations}
\begin{align}
    {^{k+1}}\Tilde{\boldsymbol{\alpha}} &= \arg\min_{\Tilde{\boldsymbol{\alpha}}} \mathcal{L}({^{k}}\boldsymbol{\xi}, \Tilde{\boldsymbol{\alpha}}, {^k}\Tilde{\boldsymbol{\beta}}, {^k}\Tilde{\mathbf{d}}, {^k}\boldsymbol{\lambda} ) = f_1({^k}\boldsymbol{\xi})\label{alpha_step}\\
    {^{k+1}}\Tilde{\boldsymbol{\beta}} &= \arg\min_{\Tilde{\boldsymbol{\beta}}} \mathcal{L}({^{k}}\boldsymbol{\xi}, {^{k+1}}\Tilde{\boldsymbol{\alpha}}, \Tilde{\boldsymbol{\beta}}, {^k}\Tilde{\mathbf{d}}, {^k}\boldsymbol{\lambda} ) = f_2({^k}\boldsymbol{\xi})\label{beta_step}\\
    {^{k+1}}\Tilde{\mathbf{d}} &= \arg\min_{\mathbf{d}_{min}\leq \Tilde{\mathbf{d}}\leq \mathbf{d}_{max}} \mathcal{L}({^{k}}\boldsymbol{\xi}, {^{k+1}}\Tilde{\boldsymbol{\alpha}}, {^{k+1}}\Tilde{\boldsymbol{\beta}}, \Tilde{\mathbf{d}}, {^k}\boldsymbol{\lambda} ) = f_3({^k}\boldsymbol{\xi})\label{d_step}\\
    {^{k+1}}\boldsymbol{\lambda} &= {^{k}}\boldsymbol{\lambda}+\rho \mathbf{F}^T (\mathbf{F}{^k}\boldsymbol{\xi}-\mathbf{e}( {^{k+1}}\Tilde{\boldsymbol{\alpha}}, {^{k+1}}\Tilde{\boldsymbol{\beta}}, {^{k+1}}\Tilde{\mathbf{d}}    )    ) \label{lambda_step} \\
    {^{k+1}}{\boldsymbol{\xi}} &= \arg\min_{\mathbf{A}\boldsymbol{\xi} = \mathbf{b}} \mathcal{L}(\boldsymbol{\xi}, \mathbf{e}( {^{k+1}}\Tilde{\boldsymbol{\alpha}}, {^{k+1}}\Tilde{\boldsymbol{\beta}}, {^{k+1}}\Tilde{\mathbf{d}}    ), {^{k+1}}\boldsymbol{\lambda} ) \label{xi_step}\\
    &= \mathbf{M}^{-1}\boldsymbol{\eta} 
\end{align}
\end{subequations}
\normalsize
\small
\begin{align}
    \textbf{M} = \begin{bmatrix}
        \mathbf{I}+\rho\mathbf{F}^T\mathbf{F} & \textbf{A}^{T} \\ 
        \mathbf{A} & \mathbf{0}
    \end{bmatrix}^{-1} \hspace{-0.19cm}, \boldsymbol{\eta} = \begin{bmatrix}
        \rho\mathbf{F}^T {^{k+1}}\mathbf{e}+{^{k+1}}\boldsymbol{\lambda}+\overline{\boldsymbol{\xi}}\\
        \mathbf{b}
    \end{bmatrix} 
\end{align}
\normalsize

The minimization \eqref{alpha_step}-\eqref{d_step} have a closed-form solution which can be expressed as a function of ${^k}\boldsymbol{\xi}$ \cite{adajania2023amswarm}, \cite{rastgar2021gpu}. Similarly, \eqref{xi_step} is simply a equality constrained QP and thus has an explicit formula for its solution. Moreover, since ${^{k+1}}\mathbf{e}$ and ${^{k+1}}\boldsymbol{\lambda}$ are explicit functions of ${^k}\boldsymbol{\xi}$, ${^k}\boldsymbol{\lambda}$, \eqref{lambda_step}-\eqref{xi_step} constitutes the fixed-point iteration $\mathbf{f}_{FP}$ presented in \eqref{fixed_point}.

A few points about the AM steps are worth noting. First, since every step has a closed-form solution, we can easily unroll them into a differentiable computational graph. Second, steps \eqref{alpha_step}-\eqref{lambda_step} do not involve any matrix factorization and only requires element-wise operation. Thus, they can be trivially batched across GPUs. Moreover, in step \eqref{xi_step}, the matrix $\mathbf{M}$ is idependent of the input $\overline{\boldsymbol{\xi}}$ sampled from the VQ-VAE/CVAE. Thus, its factorizations can be pre-stored. This also implies that the batched version of minimization \eqref{xi_step} only requires matrix-matrix products, which is easy to compute over GPU

\noindent \textbf{Primal Residual:} The primal residual vector $\mathbf{r}_p$ at $(k+1)^{th}$ iteration is given by the following.
\begin{align}
    \mathbf{r}_p = (\mathbf{F}{^{k+1}}\boldsymbol{\xi}-\mathbf{e}( {^{k+1}}\Tilde{\boldsymbol{\alpha}}, {^{k+1}}\Tilde{\boldsymbol{\beta}}, {^{k+1}}\Tilde{\mathbf{d}}    )    )
    \label{primal_residual}
\end{align}
\noindent Essentially, $\mathbf{r}_p$ dictates how well the non-convex equality constraints are satisfied. It is easy to see that $\Vert\mathbf{r}_p\Vert_2 = 0$ implies that our reformulation \eqref{inter_agent_polar}-\eqref{workspace_polar} holds and the original inter-agent \eqref{inter_robot_con} and workspace constraints \eqref{workspace_con} are satisfied.


\bibliography{references}
\bibliographystyle{IEEEtran}

\end{document}